\documentclass[10pt,twocolumn,letterpaper]{article}

\usepackage{iccv}
\usepackage{times}
\usepackage{epsfig}
\usepackage{graphicx}
\usepackage{amsmath}
\usepackage{amssymb}
\usepackage{array}
\usepackage{makecell}
\usepackage{ bbold }

\usepackage{wrapfig}
\usepackage{algpseudocode}
\usepackage{algorithm}
\usepackage{algorithmicx}

\makeatletter
\newcommand{\thickhline}{%
    \noalign {\ifnum 0=`}\fi \hrule height 2pt
    \futurelet \reserved@a \@xhline
}
\newcolumntype{"}{@{\hskip\tabcolsep\vrule width 1.5pt\hskip\tabcolsep}}
\makeatother

\newcommand{\sw}{{\bf w}}

\newcommand{\cL}{\mathcal{L}}
\newcommand{\sL}{\ell}

\usepackage[pagebackref=true,breaklinks=true,letterpaper=true,colorlinks,bookmarks=false]{hyperref}

 \iccvfinalcopy % *** Uncomment this line for the final submission

\def\iccvPaperID{166} % *** Enter the ICCV Paper ID here

\begin{document}

%%%%%%%%% TITLE
\title{Holistically-Nested Edge Detection}

\author{Saining Xie\\
Dept. of CSE and Dept. of CogSci \\
University of California, San Diego\\
9500 Gilman Drive, La Jolla, CA 92093\\
{\tt\small s9xie@eng.ucsd.edu}
\and
Zhuowen Tu\\
Dept. of CogSci and Dept. of CSE \\
University of California, San Diego\\
9500 Gilman Drive, La Jolla, CA 92093\\
{\tt\small ztu@ucsd.edu}
}

\maketitle
%\thispagestyle{empty}

%%%%%%%%% ABSTRACT
\vspace{-8mm}
\begin{abstract}
\vspace{-3mm}
  We develop a new edge detection algorithm that addresses two important issues in this long-standing vision problem: (1) holistic image training and prediction; and (2) multi-scale and multi-level feature learning. Our proposed method, holistically-nested edge detection (HED), performs image-to-image prediction by means of a deep learning model that leverages fully convolutional neural networks and deeply-supervised nets.  HED automatically learns rich hierarchical representations (guided by deep supervision on side responses) that are important in order to %approach the human ability 
resolve the challenging ambiguity in edge and object boundary detection. We significantly advance the state-of-the-art on the BSD500 dataset (ODS F-score of $.782$) and the NYU Depth dataset (ODS F-score of $.746$), and do so with an improved speed ($0.4$s per image) that is orders of magnitude faster than some recent CNN-based edge detection algorithms.
\end{abstract}

%%%%%%%%% BODY TEXT
\vspace{-3mm}
\section{Introduction}
\vspace{-2mm}

In this paper, we address the problem of detecting edges and object boundaries in natural images. This problem is both fundamental and of great importance to a variety of computer vision areas ranging from traditional tasks such as visual saliency, segmentation, object detection/recognition, tracking and motion analysis, medical imaging, structure-from-motion and 3D reconstruction, to modern applications like autonomous driving, mobile computing, and image-to-text analysis. 
%Researchers in the field have been relatively non-specific, frequently referring to both edge and object boundary detection under the heading of ``integrated edge detection''. 
%task in general. 
It has been long understood that precisely localizing edges in natural images involves visual perception of various ``levels'' \cite{hubel1962receptive,marr1980theory}. A relatively comprehensive data collection and cognitive study \cite{martin2004learning} shows that while different subjects do have somewhat different preferences regarding where to place the edges and boundaries, there was nonetheless impressive consistency between subjects, e.g. reaching F-score $0.80$ in the consistency study \cite{martin2004learning}.

\begin{figure}[!htp]
\begin{center}
   \includegraphics[width=1\linewidth]{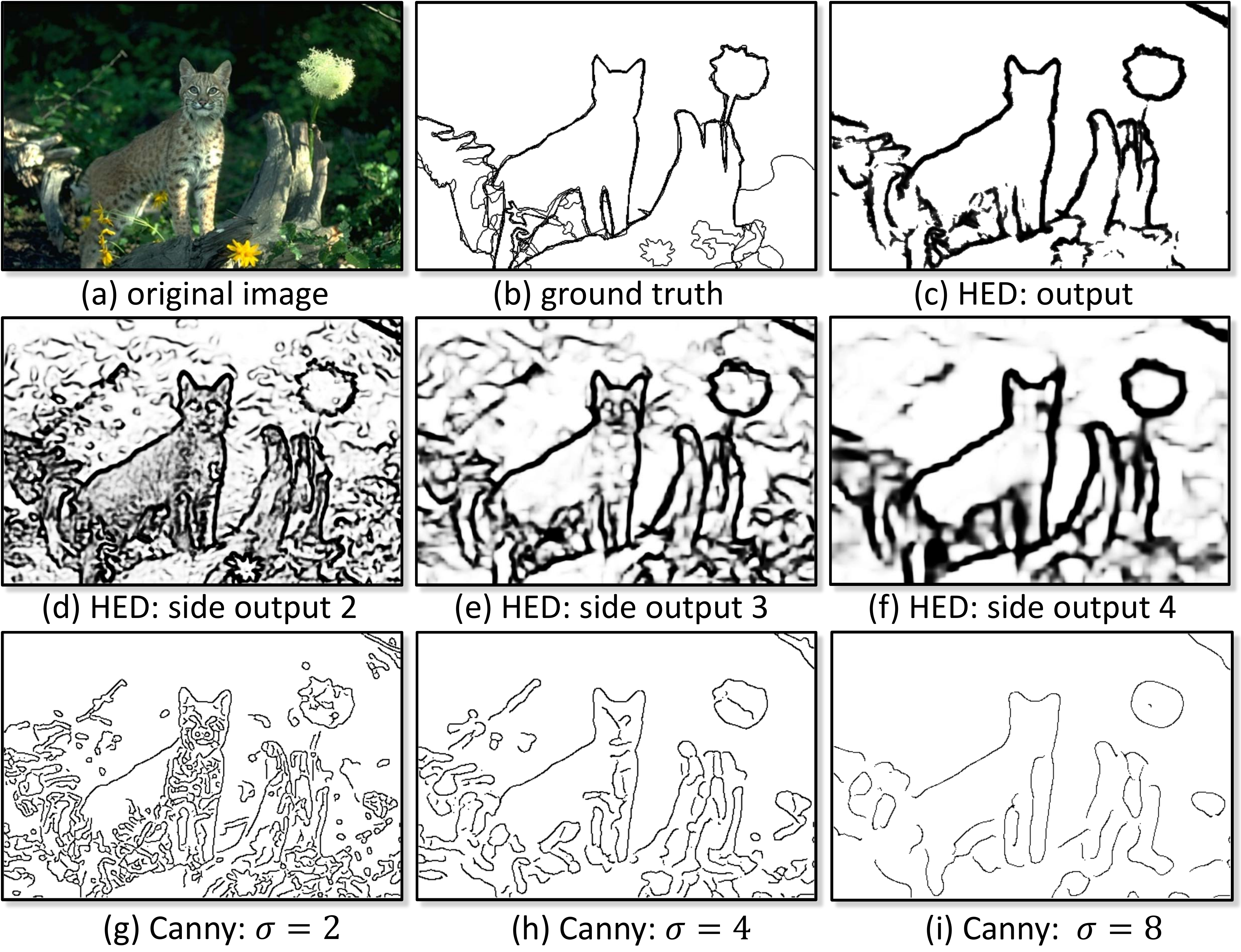}
  \caption{\footnotesize Illustration of the proposed HED algorithm. In the first row: (a) shows an example test image in the BSD500 dataset \cite{martin2004learning}; (b) shows its corresponding edges as annotated by human subjects; (c) displays the HED results. In the second row: (d), (e), and (f), respectively, show side edge responses from layers $2$, $3$, and $4$ of our convolutional neural networks. In the third row: (g), (h), and (i), respectively, show edge responses from the Canny detector \cite{canny1986computational} at the scales $\sigma=2.0$, $\sigma=4.0$, and $\sigma=8.0$. HED shows a clear advantage in consistency over Canny.  %; note that while there is a multi-scale aspect, the responses do not show nesting nor do they show direct connection. 
}
\label{fig:illustration}
\end{center}
\vspace{-10mm}
\end{figure}

The history of computational edge detection is extremely rich; we now highlight a few representative works that have proven to be of great practical importance. Broadly speaking, one may categorize works into a few groups such as I: {\em early pioneering methods} like the Sobel detector \cite{kittler1983accuracy}, zero-crossing \cite{marr1980theory,torre1986edge}, and the widely adopted Canny detector \cite{canny1986computational}; methods driven by II: {\em information theory} on top of features arrived at through careful manual design, such as Statistical Edges \cite{konishi2003statistical}, Pb \cite{martin2004learning}, and gPb \cite{arbelaez2011contour}; and III: {\em learning-based} methods that remain reliant on features of human design, such as BEL \cite{dollar2006supervised}, Multi-scale \cite{ren2008multi}, Sketch Tokens \cite{lim2013sketch}, and Structured Edges \cite{Dollar2015PAMI}. In addition, there has been a recent wave of development using {\em Convolutional Neural Networks} that emphasize the importance of automatic hierarchical feature learning, including $N^4$-Fields \cite{ganin2014n}, DeepContour \cite{shendeepcontour}, DeepEdge \cite{bertasius2014deepedge}, and CSCNN \cite{hwang2015pixel}. Prior to this explosive development in deep learning, the Structured Edges method (typically abbreviated SE) \cite{Dollar2015PAMI} emerged as one of the most celebrated systems for edge detection, thanks to its state-of-the-art performance on the BSD500 dataset \cite{martin2004learning} (with, e.g., F-score of $.746$) and its practically significant speed of $2.5$ frames per second.
%Notwithstanding this impressive performance, there nonetheless remains a big gap between SE performance and human performance.
% --- for example, SE still heavily relies on manually designed features.
Recent CNN-based methods \cite{ganin2014n,shendeepcontour,bertasius2014deepedge,hwang2015pixel} have demonstrated promising F-score performance improvements over SE. 
However, there still remains large room for improvement in these CNN-based methods, in both F-score performance and in speed --- at present, time to make a prediction ranges from several seconds \cite{ganin2014n} to a few hours \cite{bertasius2014deepedge} (even when using modern GPUs).

Here, we develop an end-to-end edge detection system, holistically-nested edge detection (HED), that automatically learns the type of rich hierarchical features that are crucial if we are to approach the human ability to resolve ambiguity in natural image edge and object boundary detection. We use the term ``holistic'', because HED, despite not explicitly modeling structured output, aims to train and predict edges in an image-to-image fashion. With ``nested'', we emphasize the inherited and progressively refined edge maps produced as side outputs --- we intend to show that the path along which each prediction is made is common to each of these edge maps, with successive edge maps being more concise.
%, in addition to feature maps generated and fed through the convolutional neural network hierarchy. 
This integrated learning of hierarchical features is in distinction to previous multi-scale approaches \cite{witkin1984scale,yuille1986scaling,ren2008multi} in which scale-space edge fields are neither automatically learned nor hierarchically connected. Figure \ref{fig:illustration} gives an illustration of an example image together with the human subject ground truth annotation, as well as results by the proposed HED edge detector (including the side responses of the individual layers), and results by the Canny edge detector \cite{canny1986computational} with different scale parameters. Not only are Canny edges at different scales not directly connected, they also exhibit spatial shift and inconsistency.
%Adopting convolutional neural networks \cite{ganin2014n,shendeepcontour,bertasius2014deepedge,hwang2015pixel}, an ``overall'' edge result is produced but only guided at the very top layer. In addition, these CNN-based methods are based on patch-to-patch training and testing, leading to a speed disadvantage. % relative to alternative approaches \cite{canny1986computational,Dollar2015PAMI}. 

The proposed holistically-nested edge detector (HED) tackles two critical issues: (1) holistic image training and prediction, inspired by fully convolutional neural networks \cite{long2014fully}, for image-to-image classification (the system takes an image as input, and directly produces the edge map image as output); 
 and (2) nested multi-scale feature learning, inspired by  deeply-supervised nets \cite{DSN}, that performs deep layer supervision to ``guide'' early classification results. We find that the favorable characteristics of these underlying techniques manifest in HED being both accurate and computationally efficient. 
%HED is, as a result, an end-to-end system that significantly advances the state-of-the-art on the BSD500 dataset (ODS F-score of $0.782$) and the NYU Depth dataset (ODS F-score of $0.746$), and that is applicable in practice owing to an improved speed ($0.4$ second per image) that is orders of magnitude faster than recent CNN-based edge detection algorithms.
\begin{figure*}[!htp]
\begin{center}
  %\fbox{\includegraphics[width=1\linewidth]{multiscale_gray.pdf}}
\includegraphics[width=0.85\linewidth]{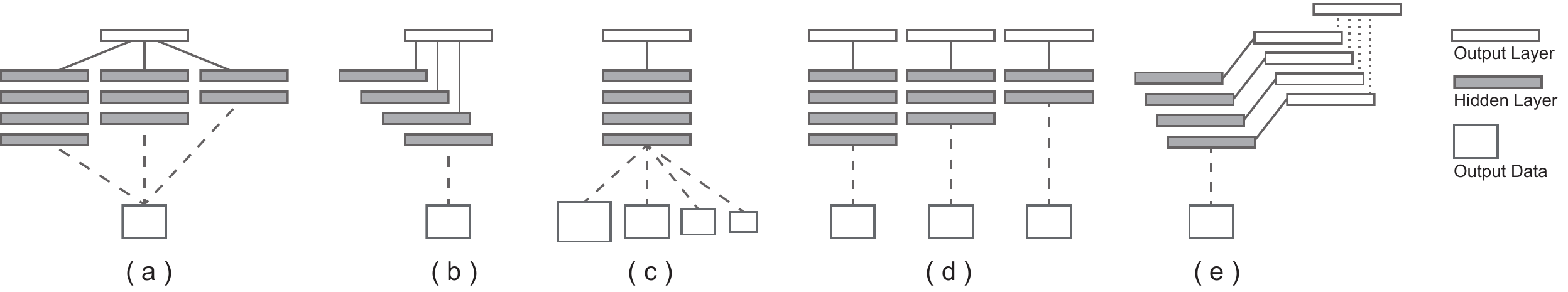}
  \caption{\footnotesize Illustration of different multi-scale deep learning architecture configurations: (a) {multi-stream} architecture; (b) {skip-layer net} architecture; (c) a single model running on multi-scale inputs; (d) separate training of different networks; (e) our proposed {holistically-nested} architectures, where multiple side outputs are added.}
  \label{fig:multiscale}
\end{center}
\vspace{-0.8cm}
\end{figure*}

\vspace{-2mm}
\section{Holistically-Nested Edge Detection}
\vspace{-1mm}
In this section, we describe in detail the formulation of our proposed edge detection system. We start by discussing related neural-network-based approaches, particularly those that emphasize multi-scale and multi-level feature learning.
The task of edge and object boundary detection is inherently challenging. After decades of research, there have emerged a number of properties that are key and that are likely to play a role in a successful system: (1) carefully designed and/or learned features \cite{martin2004learning,dollar2006supervised}, (2) multi-scale response fusion \cite{witkin1984scale,ruderman1994statistics,ren2008multi}, (3) engagement of different levels of visual perception \cite{hubel1962receptive,marr1980theory,van1994neural,hou2013boundary} such as mid-level Gestalt law information \cite{elder2002ecological}, (4) incorporating structural information (intrinsic correlation carried within the input data and output solution) \cite{Dollar2015PAMI} and context (both short- and long- range interactions) \cite{tu2008auto}, (5) making holistic image predictions (referring to approaches that perform prediction by taking the image contents globally and directly) \cite{liu2011nonparametric}, (6) exploiting 3D geometry \cite{hoiem2008putting}, and (7) addressing occlusion boundaries \cite{hoiem2007recovering}. 

\begin{figure}[!htp]
	\begin{center}
		\includegraphics[width=1\linewidth]{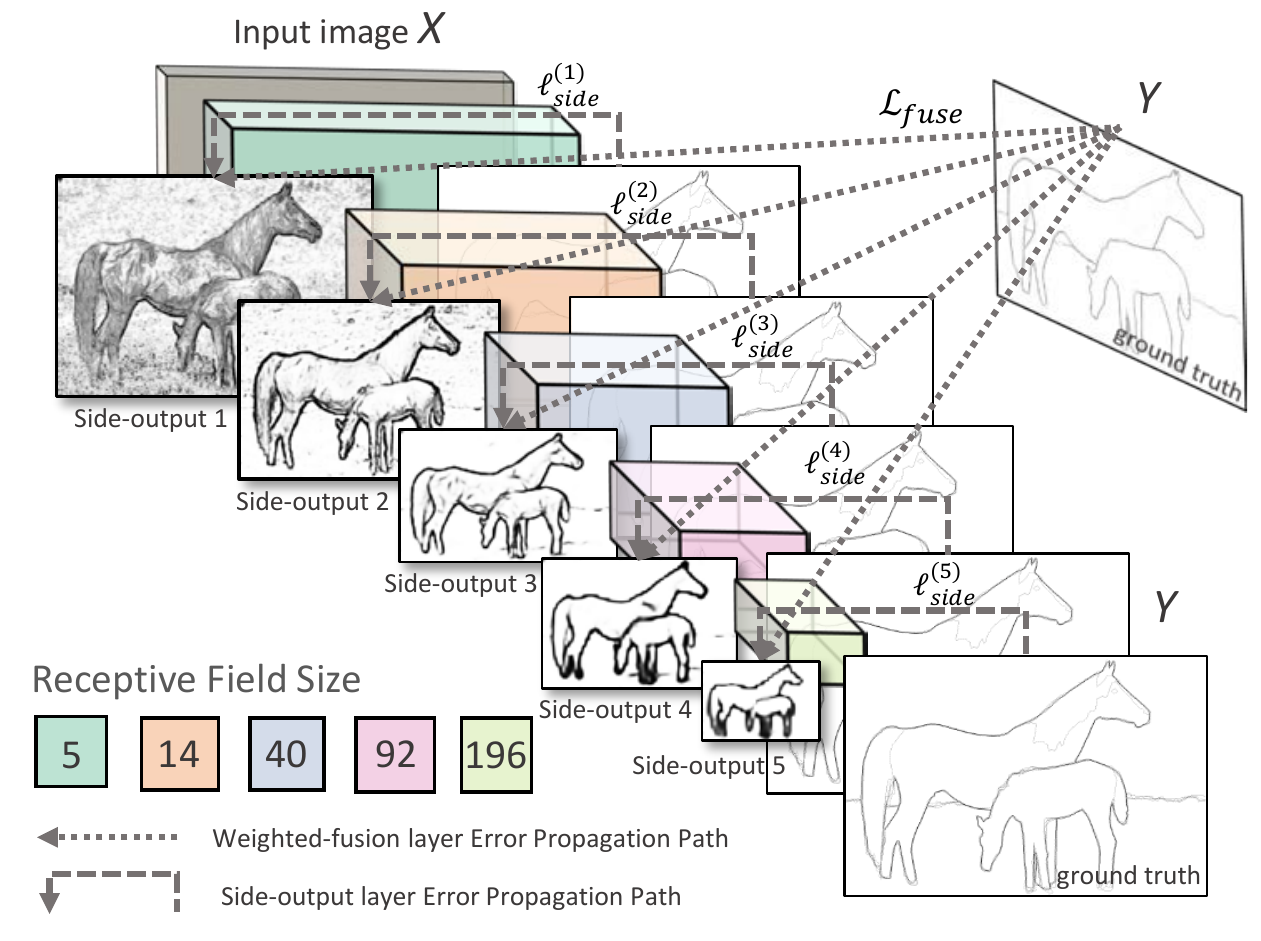}
	\end{center}
	\caption{\footnotesize Illustration of our network architecture for edge detection, highlighting the error backpropagation paths. Side-output layers are inserted after convolutional layers. Deep supervision is imposed at each side-output layer, guiding the side-outputs towards edge predictions with the characteristics we desire. The outputs of HED are multi-scale and multi-level, with the side-output-plane size becoming smaller and the receptive field size becoming larger. One weighted-fusion layer is added to automatically learn how to combine outputs from multiple scales. The entire network is trained with multiple error propagation paths (dashed lines). }
	\vspace{-3mm}
	\label{fig:network}
\end{figure}

Structured Edges (SE) \cite{Dollar2015PAMI} primarily focuses on three of these aspects: using a large number of manually designed features (property 1), fusing multi-scale responses (property 2), and incorporating structural information (property 4). A recent wave of work using CNN for patch-based edge prediction \cite{ganin2014n,shendeepcontour,bertasius2014deepedge,hwang2015pixel} contains an alternative common thread that focuses on three aspects: automatic feature learning (property 1), multi-scale response fusion (property 2), and possible engagement of different levels of visual perception (property 3).
However, due to the lack of deep supervision (that we include in our method), the multi-scale responses produced at the hidden layers in \cite{bertasius2014deepedge,hwang2015pixel} are less semantically meaningful, since feedback must be back-propagated through the intermediate layers. 
More importantly, their patch-to-pixel or patch-to-patch strategy results in significantly downgraded training and prediction efficiency. 
By ``holistically-nested'', we intend to emphasize that we are producing an end-to-end edge detection system, a strategy inspired by fully convolutional neural networks \cite{long2014fully}, but with additional deep supervision on top of trimmed VGG nets \cite{vgg} (shown in Figure \ref{fig:network}).
%; in this system , we not only train image-to-image prediction but also include hidden layers that are supervised to generate ``side'' edge maps. 
%that are directly connected through layers of feature responses
%Our proposed network (shown in Figure \ref{fig:network}), instead pursues holistic image-to-image training and prediction,
In the absence of deep supervision and side outputs, a fully convolutional network \cite{long2014fully} (FCN) produces a less satisfactory result (e.g. F-score $.745$ on BSD500) than HED, since edge detection demands highly accurate edge pixel localization.
One thing worth mentioning is that our image-to-image training and prediction strategy still has not explicitly engaged contextual information, since constraints on the neighboring pixel labels are not directly enforced in HED. In addition to the speed gain over patch-based CNN edge detection methods, the performance gain is largely due to three aspects: (1) FCN-like image-to-image training allows us to simultaneously train on a significantly larger amount of samples (see Table \ref{tb:bsds}); (2) deep supervision in our model guides the learning of more transparent features (see Table \ref{tab:hed_fcn}); (3) interpolating the side outputs in the end-to-end learning encourages coherent contributions from each layer (see Table \ref{tab:hed_var}).

\vspace{-2mm}
\subsection{Existing multi-scale and multi-level NN}
\vspace{-2mm}
%Multi-scale or hierarchical processing is an important learning strategy. However, traditional feed-forward neural networks cannot explicitly takes advantages of multi-scale learning, since the decision made at the final output layer m are purely based on the directly connected hidden layer m-1. Recently people notice that lower level features learned in the neural networks are complementary to the high level features, and combining those two type of features can be helpful.

Due to the nature of hierarchical learning in the deep convolutional neural networks, the concept of multi-scale and multi-level learning might differ from situation to situation. For example, multi-scale learning can be ``inside'' the neural network, in the form of increasingly larger receptive fields and downsampled (strided) layers. In this ``inside'' case, the feature representations learned in each layer are naturally multi-scale. On the other hand, multi-scale learning can be ``outside'' of the neural network, for example by ``tweaking the scales'' of input images. While these two variants have some notable similarities, we have seen both of them applied to various tasks.

We continue by next formalizing the possible configurations of multi-scale deep learning into four categories, namely, {\em multi-stream} learning, {\em skip-net} learning, a {\em single model} running on multiple inputs, and training of {\em independent} networks. An illustration is shown in Fig~\ref{fig:multiscale}. Having these possibilities in mind will help make clearer the ways in which our proposed {\em holistically-nested} network approach differs from previous efforts and will help to highlight the important benefits in terms of representation and efficiency.

{\em Multi-stream learning} \cite{buyssens2013multiscale,neverova2014multi} A typical multi-stream learning architecture is illustrated in Fig~\ref{fig:multiscale}(a). Note that the multiple (parallel) network streams have different parameter numbers and receptive field sizes, corresponding to multiple scales. Input data are simultaneously fed into multiple streams, after which the concatenated feature responses produced by the various streams are fed into a global output layer to produce the final result. 

{\em Skip-layer network learning:} Examples of this form of network include \cite{long2014fully,hariharan2014hypercolumns,bertasius2014deepedge,sermanet2012convolutional,ganin2014n}. The key concept in ``skip-layer'' network learning is shown in Fig~\ref{fig:multiscale}(b). Instead of training multiple parallel streams, the topology for the skip-net architecture centers on a primary stream. Links are added to incorporate the feature responses from different levels of the primary network stream, and these responses are then combined in a shared output layer.

A common point in the two settings above is that, in both of the architectures, there is only one output loss function with a single prediction produced. However, in edge detection, it is often favorable (and indeed prevalent) to obtain multiple predictions to combine the edge maps together.
%For image-level classification tasks, the result produced by this output layer is a single label. If we reinterpret the convolutional networks for pixel-wise classification tasks, e.g. edge detection, the result is a presumably finer-scale edge map. Nevertheless we got 

%We note that the concept of multi-scale learning also differs in ``feature'' and ``prediction''. In this paper, we are focusing on pixel-wise edge learning problem, where we put more emphasize on the aspect of multi-scale predictions.

% Augmentation and ensemble
{\em Single model on multiple inputs:} To get multi-scale predictions, one can also run a single network (or networks with tied weights) on multiple (scaled) input images, as illustrated in Fig~\ref{fig:multiscale}(c). This strategy can happen at both the training stage (as data augmentation) and at the testing stage (as ``ensemble testing''). One notable example is the tied-weight pyramid networks \cite{farabet2013learning}. 
%In training stage, people can augment the data by adding resized images. During inference, one can run the single model on images from multiple scales, resize the prediction to the original image size and average the predictions together. 
This approach is also common in non-deep-learning based methods \cite{Dollar2015PAMI}. Note that ensemble testing impairs the prediction efficiency of learning systems, especially with deeper models\cite{bertasius2014deepedge,ganin2014n}.

{\em Training independent networks:} As an extreme variant to Fig~\ref{fig:multiscale}(a), one might pursue Fig~\ref{fig:multiscale}(d), in which multi-scale predictions are made by training multiple independent networks with different depths and different output loss layers. This might be practically challenging to implement as this duplication would multiply the amount of resources required for training. 

{\em Holistically-nested networks:} We list these variants to help clarify the distinction between existing approaches and our proposed holistically-nested network approach, illustrated in Fig~\ref{fig:multiscale}(e). There is often significant redundancy in existing approaches, in terms of both representation and computational complexity. Our proposed holistically-nested network is a relatively simple variant that is able to produce predictions from multiple scales. The architecture can be interpreted as a ``holistically-nested'' version of the ``independent networks'' approach in Fig~\ref{fig:multiscale}(d), motivating our choice of name. Our architecture comprises a single-stream deep network with multiple side outputs. This architecture resembles several previous works, particularly the deeply-supervised net\cite{DSN} approach in which the authors show that hidden layer supervision can improve both optimization and generalization for image classification tasks. The multiple side outputs also give us the flexibility to add an additional fusion layer if a unified output is desired.

\subsection{Formulation}
Here we formulate our approach for edge prediction. 
\noindent \textbf{Training Phase} 
We denote our input training data set by $S=\{(X_{n},Y_{n}),n=1,\dots,N\}$, where sample $X_{n}=\{x_j^{(n)}, j=1,\ldots,|X_{n}|\}$ denotes the raw input image and $Y_{n}=\{y_j^{(n)}, j=1,\ldots,|X_{n}|\}, y_j^{(n)}\in\{0, 1\}$ denotes the corresponding ground truth binary edge map for image $X_{n}$. We subsequently drop the subscript $n$ for notational simplicity, since we consider each image holistically and independently. Our goal is to have a network that learns features from which it is possible to produce edge maps approaching the ground truth. For simplicity, we denote the collection of all standard network layer parameters as $\bf{W}$. Suppose in the network we have $M$ side-output layers. Each side-output layer is also associated with a classifier, in which the corresponding weights are denoted as $\sw=(\sw^{(1)},\ldots,\sw^{(M)})$.  We consider the objective function 
\begin{equation}
\vspace{-3mm}
\cL_{\text{side}}({\bf W},\sw) = \sum_{m=1}^{M}\alpha_{m}\sL_{\text{side}}^{(m)}({\bf W},{\bf w}^{(m)}),
\vspace{-0mm}
\end{equation}
where $\sL_{\text{side}}$ denotes the image-level loss function for side-outputs.
%In addition, we denote $\hat{\bf G}^{(m)}$ as the (predicted) edge map produced by side-output layer $m$, upsampled to original size when necessary, 
%$\Delta$ is an ``energy function'', (e.g. cross-entropy) computing the loss of the predicted edge map over the ground truth target, and $\alpha_m$ is a hyper-parameter controlling the loss weight for each individual side-output layer.
In our image-to-image training, the loss function is computed over all pixels in a training image $X=(x_j, j=1,\ldots, |X|)$ and edge map $Y=(y_j, j=1,\ldots, |X|), y_j \in \{0, 1\}$. For a typical natural image, the distribution of edge/non-edge pixels is heavily biased: 90\% of the ground truth is non-edge. A cost-sensitive loss function is proposed in \cite{hwang2015pixel}, with additional trade-off parameters introduced for biased sampling. 

We instead use a simpler strategy to automatically balance the loss between positive/negative classes. We introduce a class-balancing weight $\beta$ on a per-pixel term basis. Index $j$ is over the image spatial dimensions of image $X$. Then we use this class-balancing weight as a simple way to offset this imbalance between edge and non-edge. Specifically, we define the following class-balanced cross-entropy loss function used in Equation (1) 
%Fixing an image sample ${\bf I}$ and a side-output layer. 
%\begin{equation}
%\begin{split}
%\sL_{\text{side}}^{(m)}& ({\bf W}, {\bf w}^{(m)})   = \\ &\sum_{j = 1}^{|Y|} (-\beta y_j\log p_j^{(m)} - (1-\beta) (1 - y_j) \log p_j^{(m)})
%\end{split}
%\end{equation}
% where $p_j^{(m)} = Pr(y_j=1|X; {\bf W}, {\bf w}^{(m)}) = \sigma(a_j^{(m)})\in[0,1]$ is computed using sigmoid function $\sigma(.)$  and  $(a^{(m)}_j, j = 1,\dots, |Y|)$ are the activations of the side-output $m$, 
% $\beta = \frac{|{Y_-}|}{|Y|}$ and $1-\beta$ = $\frac{|Y_+|}{|Y|}$ where ${|{Y_-}|}$ and ${|{Y_+}|}$ denotes the edge/non-edge ground truth labeling set respectively. 

\vspace{-4mm}
\[
\sL_{\text{side}}^{(m)} ({\bf W}, {\bf w}^{(m)})   = -\beta\sum_{j \in Y_+}\log \operatorname{Pr}(y_j = 1|X; {\bf W}, {\bf w}^{(m)}) 
\]
\vspace{-4mm}
\begin{equation}
-(1-\beta)\sum_{j \in Y_-} \log \operatorname{Pr}(y_j = 0| X; {\bf W}, {\bf w}^{(m)})
\end{equation}
 where $\beta = |{Y_-}|/|Y|$ and $1-\beta$ = $|Y_+|/|Y|$. ${|{Y_-}|}$ and ${|{Y_+}|}$ denote the edge and non-edge ground truth label sets, respectively. $\operatorname{Pr}(y_j=1|X; {\bf W}, {\bf w}^{(m)}) = \sigma(a_j^{(m)})\in[0,1]$ is computed using sigmoid function $\sigma(.)$ on the activation value at pixel $j$.
 At each side output layer, we then obtain edge map predictions
 $\hat{Y}_{\text{side}}^{(m)} = \sigma(\hat{A}_{\text{side}}^{(m)})$, where $\hat{A}_{\text{side}}^{(m)} \equiv \{a^{(m)}_j,\  j=1,\dots,|Y|\}$ are activations of the side-output of layer $m$.
 
%\noindent \textbf{Weighted-fusion layer} 
%One natural question is how to best utilize the side-outputs produced by our framework; while the most straightforward approach is to simply average all the edge map predictions, one can also learn a classifier to optimize the combination weights, either at pixel level (locally connected) or at image level. %The desire for a practically useful edge detector leads us to favor a system that is completely end-to-end, motivating us to add a ``weighted-fusion'' layer to the network.
To directly utilize side-output predictions, we add a ``weighted-fusion'' layer to the network and (simultaneously) learn the fusion weight during training. Our loss function at the fusion layer $\cL_{\text{fuse}}$ becomes 
%\[
%\cL_{\text{fuse}}({\bf W},\sw,{\bf h}) = Dist(Y, \;\;\sum_{m=1}^M h_m \hat{Y}^{(m)})
%\]
%where $\hat{Y}^{(m)} \equiv \{Pr(y_j = 1|X; {\bf W}, {\bf w}^{(m)}), j=1,\dots,|Y|\} $ is the edge map probability predictions on side output layer $m$ and ${\bf h}=(h_1,\dots,h_M)$ is the fusion weight. The distance between fused predictions and ground truth label map is still optmized with cross-entropy loss. The overall loss function then becomes
\begin{equation}
\label{eqn:fuse}
\cL_{\text{fuse}}({\bf W},\sw,{\bf h}) = \operatorname{Dist}(Y,\ \hat{Y}_{\text{fuse}})
\end{equation}
where $\hat{Y}_{\text{fuse}} \equiv \sigma(\sum_{m=1}^M h_m\hat{A}_{\text{side}}^{(m)})$ where ${\bf h}=(h_1,\dots,h_M)$ is the fusion weight. $\operatorname{Dist}(\cdot,\cdot)$ is the distance between the fused predictions and the ground truth label map, which we set to be cross-entropy loss. Putting everything together, we minimize the following objective function via standard (back-propagation) stochastic gradient descent:
\begin{equation}
\label{eqn:all}
(\bf{W},\sw,{\bf h})^{\star} = \arg\!\min (\cL_{\text{side}}(\bf{W},\sw) +  \cL_{\text{fuse}}(\bf{W},\sw,{\bf h}))
\end{equation}
See section~\ref{sec:exp} for detailed hyper-parameter and experiment settings.

\noindent \textbf{Testing phase} During testing, given image $X$, we obtain edge map predictions from both the side output layers and the weighted-fusion layer:
\begin{equation}
(\hat{Y}_{\text{fuse}},\ \hat{Y}_{\text{side}}^{(1)},\ \dots,\ \hat{Y}_{\text{side}}^{(M)}) = \operatorname{CNN}(X, (\bf{W},\sw,{\bf h})^{\star}),
\end{equation}
where $\operatorname{CNN}(\cdot)$ denotes the edge maps produced by our network. The final unified output can be obtained by further aggregating these generated edge maps. The details will be discussed in section~\ref{sec:exp}.
\begin{equation}
\hat{Y}_{\text{HED}} = \operatorname{Average}(\hat{Y}_{\text{fuse}},\ \hat{Y}_{\text{side}}^{(1)},\ \dots ,\ \hat{Y}_{\text{side}}^{(M)})
\end{equation}

%We add a ``weighted-fusion'' layer to the network and link all the side-output layer predictions and (simultaneously) learns the fusion weight during training. Our loss function at the fusion layer $\cL_{\text{fuse}}$ becomes 
%\[
%\cL_{\text{fuse}}({\bf W},\sw,{\bf h}) = Dist(Y, \;\;\sum_{m=1}^M h_m \hat{Y}^{(m)})
%\]
%where $\hat{Y}^{(m)}={\bf Pr}({\bf 1}|X; {\bf W},\sw^{(m)})\equiv Pr(y_j = 1|X; {\bf W}, {\bf w}^{(m)}), j=1,\dots,|Y| $ is the edge map probability predictions on side output layer $m$ and ${\bf h}=(h_1,\dots,h_M)$ is the fusion weight. The distance between fused predictions and ground truth label map is still optmized with cross-entropy loss. The overall loss function then becomes
%\begin{equation*}
%\cL(\bf{W},\sw,{\bf h}) = \cL_{\text{side}}(\bf{W},\sw) +  \cL_{\text{fuse}}(\bf{W},\sw,{\bf h})
%\end{equation*}
\section{Network Architecture}
\vspace{-2mm}
Next, we describe the network architecture of HED.
\vspace{-1mm}
\subsection{Trimmed network for edge detection}
\vspace{-1mm}
The choice of hierarchy for our framework deserves some thought. We need the architecture (1) to be deep, so as to efficiently generate perceptually multi-level features; and (2) to have multiple stages with different strides, so as to capture the inherent scales of edge maps. We must also keep in mind the potential difficulty in training such deep neural networks with multiple stages when starting from scratch.
Recently, VGGNet \cite{vgg} has been seen to achieve state-of-the-art performance in the ImageNet challenge, with great depth (16 convolutional layers), great density (stride-1 convolutional kernels), and multiple stages (five 2-stride downsampling layers). Recent work \cite{bertasius2014deepedge} also demonstrates that fine-tuning deep neural networks pre-trained on the general image classification task is useful to the low-level edge detection task. We therefore adopt the VGGNet architecture but make the following modifications: (a) we connect our side output layer to the last convolutional layer in each stage, respectively conv1\_2, conv2\_2, conv3\_3, conv4\_3, conv5\_3. The receptive field size of each of these convolutional layers is identical to the corresponding side-output layer; (b) we cut the last stage of VGGNet, including the 5th pooling layer and all the fully connected layers. The reason for ``trimming'' the VGGNet is two-fold. First, because we are expecting meaningful side outputs with different scales, a layer with stride 32 yields a too-small output plane with the consequence that the interpolated prediction map will be too fuzzy to utilize. Second, the fully connected layers (even when recast as convolutions) are computationally intensive, so that trimming layers from pool5 on can significantly reduce the memory/time cost during both training and testing. Our final HED network architecture has 5 stages, with strides 1, 2, 4, 8 and 16, respectively, and with different receptive field sizes, all nested in the VGGNet. See Table~\ref{tab:vgg} for a summary of the configurations of the receptive fields and strides.

\begin{table}[!htp]
\vspace{-3mm}
\begin{center}
\caption{\footnotesize The receptive field and stride size in VGGNet~\cite{vgg} used in HED. The bolded convolutional layers are linked to additional side-output layers.}
\label{tab:vgg}
\begin{tabular}{c|ccccc}
\hline
layer   & \textbf{c1\_2} & p1             & \textbf{c2\_2} & p2             & \textbf{c3\_3} \\ \hline
rf size & \textbf{5}     & 6              & \textbf{14}    & 16             & \textbf{40}    \\ \hline
stride  & \textbf{1}     & 2              & \textbf{2}     & 4              & \textbf{4}     \\ \hline
layer   & p3             & \textbf{c4\_3} & p4             & \textbf{c5\_3} & p5             \\ \hline
rf size & 44             & \textbf{92}    & 100            & \textbf{196}   & 212            \\ \hline
stride  & 8              & \textbf{8}     & 16             & \textbf{16}    & 32             \\ \hline
\end{tabular}
\end{center}
\vspace{-8mm}
\end{table}

%\ell = &-\frac{1}{m}( \frac{N_-^{(i)}}{N^{(i)}}\sum_{i=1}^m \sum_{j=1}^{N^{(i)}}1(y^{(i)} = 1)\log p_j^{(i)}\\
 %       & + \frac{N_+^{(i)}}{N^{(i)}}\sum_{i=1}^m \sum_{j=1}^{N^{(i)}}1(y^{(i)} = 0)\log p_j^{(i)})

\vspace{-2mm}
\subsection{Architecture alternatives}
\label{sec:alt}
Below we discuss some possible alternatives in architecture design, and in particular, the role of deep supervision of HED for the edge detection task.
\vspace{-2mm}

\begin{table}[!htp]
	\begin{center}
		\caption{\footnotesize Performance of alternative architectures on BSDS dataset. The ``fusion-output without deep supervision'' result is learned w.r.t Eqn.~\ref{eqn:fuse}. The ``fusion-output with deep supervision'' result is learned w.r.t. to Eqn.~\ref{eqn:all}.}
		\label{tab:hed_fcn}
		\begin{tabular}{  m{13.8em} | m{0.5cm} m{0.5cm} m{0.5cm}  } 
		%\hline
			\  & ODS & OIS & AP \\
			\thickhline
			%HED &  {.782} & {.802} & {.833} \\ 
			\small{FCN-8S} &  {.697} & {.715} & {.673} \\ 
			\small{FCN-2S} &  {.738} & {.756} & {.717} \\ 
			\hline
			\small{Fusion-output (w/o deep supervision)} & {.771} & {.785} & {.738} \\
			\small{Fusion-output (with deep supervision)} & {.782} & {.802} & {.787} \\
			%\hline
		\end{tabular}
	\end{center}		
	\vspace{-5mm}
\end{table}

\begin{figure}[!htp]
	\vspace{-3mm}
	\begin{center}
		\includegraphics[width=1\linewidth]{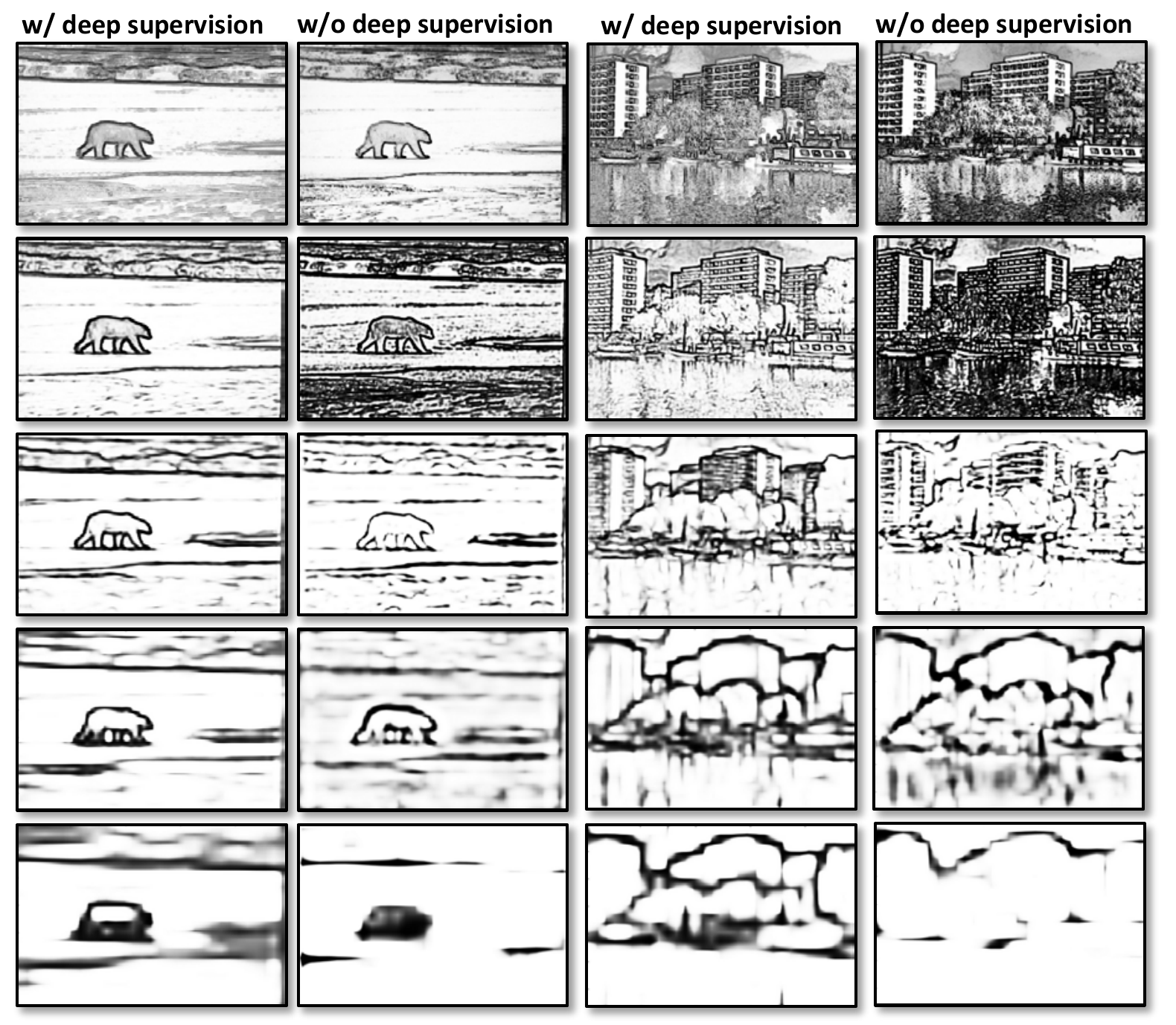}
		\caption{\footnotesize Two examples illustrating how deep supervision helps side-output layers to produce multi-scale dense predictions. Note that in the left column, the side outputs become progressively coarser and more ``global'', while critical object boundaries are preserved. In the right column, the predictions tends to lack any discernible order (e.g. in layers 1 and 2), and many boundaries are lost in later stages.}
		\vspace{-1mm}
		\label{fig:dsn}
	\end{center}
	\vspace{-9mm}
\end{figure}
\noindent \textbf{FCN and skip-layer architecture}
The topology used in the FCN model differs from that in our HED model in several aspects. As we have discussed, while FCN reinterprets classification nets for per-pixel prediction, it has only one output loss function. Thus, in FCN, although the skip net structure is a DAG that combines coarse, high-layer information with fine low-layer information, it does not explicitly produce multi-scale output predictions. We explore how this architecture can be used for the edge detection task under the same experimental setting as our HED model. We first try to directly apply the FCN-8s model by replacing the loss function with cross-entropy loss for edge detection. The results shown in first row of Table~\ref{tab:hed_fcn} are unsatisfactory, which is expected since this architecture is still not fine enough. We further explore whether the performance can be improved by adding even more links from low-level layers. We then create an FCN-2s network that adds additional links from the pool1 and pool2 layers. Still, directly applying the FCN skip-net topology falls behind our proposed HED architecture (see second row of Table~\ref{tab:hed_fcn}). With heavy tweaking of FCN, there is a possibility that one might be able to achieve competitive performance on edge detection, but the multi-scale side-outputs in HED are seen to be natural and intuitive for edge detection.

\noindent \textbf{The role of deep supervision}
Since we incorporate a weighted-fusion output layer that connects each side-output layer, there is a need to justify the adoption of the deep supervision terms (specifically, $ \sL_{\text{side}}({\bf W},{\bf \sw}^{(m)}$): now the entire network is path-connected and the output-layer parameters can be updated by back-propagation through the weighted-fusion layer error propagation path (subject to Equation~\ref{eqn:fuse}).
Here we show that deep supervision is important to obtain desired edge maps. The key characteristic of our proposed network is that each network layer is supposed to play a role as a singleton network responsible for producing an edge map at a certain scale. Here are some qualitative results based on the two variants discussed above: (1) training with both weighted-fusion supervision and deep supervision, and (2) training with weighted-fusion supervision only. We observe that with deep supervision, the nested side-outputs are natural and intuitive, insofar as the successive edge map predictions are progressively coarse-to-fine, local-to-global. On the other hand, training with only the weighted-fusion output loss gives edge predictions that lack such discernible order: many critical edges are absent at the higher layer side output; under exactly same experimental setup, the result on the benchmark dataset (row three of Table~\ref{tab:hed_fcn}) differs only marginally in F-score but displays severely degenerated average precision; without direct control and guidance across multiple scales, this network is heavily biased towards learning large structure edges.
%As we have discussed in previous section, the deep supervision plays an important role in HED. We test the model trained with/without deep supervision imposed. The model trained without deep supervision consistently performs worse (At least .01 in F-score and .02 in AP). %(fusion layer output ODS=0.771 vs. 0.782 AP=0.738 vs 0.787, averaging side outputs 1-5 ODS=0.749 vs 0.774, AP= 0.801 vs 0.822). 

\section{Experiments}
\label{sec:exp}
In this section we discuss our detailed implementation and report the performance of our proposed algorithm. 

\subsection{Implementation}
We implement our framework using the publicly available \emph{Caffe} Library and build on top of the publicly available implementations of FCN\cite{long2014fully} and DSN\cite{DSN}. Thus, relatively little engineering hacking is required. In our HED system, the whole network is fine-tuned from an initialization with the pre-trained VGG-16 Net model.

\noindent \textbf{Model parameters}
In contrast to fine-tuning CNN to perform image classification or semantic segmentation, adapting CNN to perform low-level edge detection requires special care. Differences in data distribution, ground truth distribution, and loss function all contribute to difficulties in network convergence, even with the initialization of a pre-trained model.
We first use a validation set and follow the evaluation strategy used in \cite{Dollar2015PAMI} to tune the deep model hyper-parameters. The hyper-parameters (and the values we choose) include: mini-batch size (10), learning rate (1e-6), loss-weight $\alpha_m$ for each side-output layer (1), momentum (0.9), initialization of the nested filters (0), initialization of the fusion layer weights (1/5), weight decay (0.0002), number of training iterations (10,000; divide learning rate by 10 after 5,000).
%Note that exhaustively tuning these parameters is unnecessary (thanks to our end-to-end system without any feature engineering). 
We focus on the convergence behavior of the network. We observe that whenever training converges, the deviations in F-score on the validation set tend to be very small. In order to investigate whether including additional nonlinearity helps, we also consider a setting in which we add an additional layer (with 50 filters and a ReLU) before each side-output layer; we find that this worsens performance.  On another note, we observe that our nested multi-scale framework is insensitive to input image scales; during our training process, we take advantage of this by resizing all the images to $400\times400$ to reduce GPU memory usage and to take advantage of efficient batch processing.
In the experiments that follow, we fix the values of all hyper-parameters discussed above to explore the benefits of possible variants of HED.

\noindent\textbf{Consensus sampling}
%Training the holistically nested network end-to-end is not trivial. 
In our approach, we duplicate the ground truth at each side-output layer and resize the (downsampled) side output to its original scale. Thus, there exists a mismatch in the high-level side-outputs: the edge predictions are coarse and global, while the ground truth still contains many weak edges that could even be considered as noise. This issue leads to problematic convergence behavior, even with the help of a pre-trained model. We observe that this mismatch leads to back-propagated gradients that explode at the high-level side-output layers. 
We therefore adjust how we make use of the ground truth labels in the BSDS dataset to combat this issue. Specifically, the ground truth labels are provided by multiple annotators and thus, implicitly, greater labeler consensus indicates stronger ground truth edges. We adopt a relatively brute-force solution: only assign a pixel a positive label if it is labeled as positive by at least three annotators; regard all other labeled pixels as negatives. This helps with the problem of gradient explosion in high level side-output layers. For low level layers, this consensus approach brings additional robustness to edge classification and prevents the network from being distracted by weak edges. Although not fully explored in our paper, a careful handling of consensus levels of ground truth edges might lead to further improvement. 

\noindent \textbf{Data augmentation}
Data augmentation has proven to be a crucial technique in deep networks. We rotate the images to 16 different angles and crop the largest rectangle in the rotated image; we also flip the image at each angle, leading to an augmented training set that is a factor of 32 larger than the unaugmented set.
% We find that augmenting the data to different scales is unnecessary.
During testing we operate on an input image at its original size. We also note that ``ensemble testing''  (making predictions on rotated/flipped images and averaging the predictions) yields no improvements in F-score, nor in average precision.

\noindent \textbf{Different pooling functions} Previous work \cite{bertasius2014deepedge} suggests that different pooling functions can have a major impact on edge detection results. We conduct a controlled experiment in which all pooling layers are replaced by average pooling. %In contrast to the observation in \cite{bertasius2014deepedge},
We find that using average pooling decrease the performance to ODS=.741.

\noindent \textbf{In-network bilinear interpolation}
Side-output prediction upsampling is implemented with in-network deconvolutional layers, similar to those in  \cite{long2014fully}. We fix all the deconvolutional layers to perform linear interpolation. Although it was pointed out in \cite{long2014fully} that one can learn arbitrary interpolation functions, we find that learned deconvolutions provide no noticeable improvements in our experiments. 

\noindent\textbf{Running time} 
Training takes about 7 hours on a single NVIDIA K40 GPU. For a $320 \times 480$ image, it takes HED 400 ms to produce the final edge map (including the interface overhead), which is significantly faster than existing CNN-based methods \cite{shendeepcontour,bertasius2014deepedge}.
Some previous edge detectors also try to improve performance by the less desirable expedient of sacrificing efficiency (for example, by testing on input images from multiple scales and averaging the results).

\subsection{BSDS500 dataset} 
\begin{figure}[!htp]
\vspace{-3mm}
\begin{center}
   \includegraphics[width=0.8\linewidth]{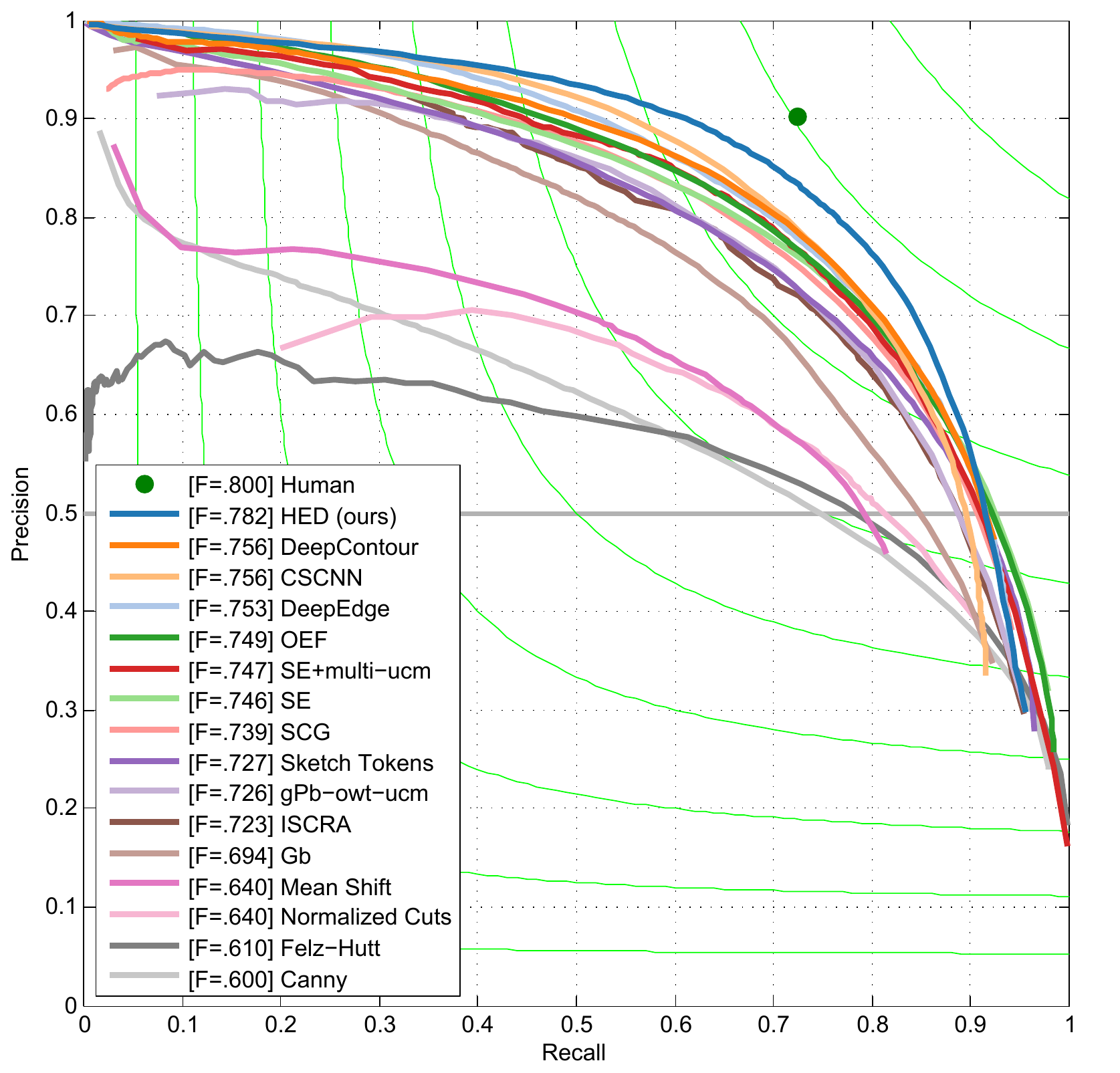}
\end{center}
\vspace{-4mm}
   \caption{\footnotesize Results on the BSDS500 dataset. Our proposed HED framework achieves the best result (ODS=.782). Compared to several recent CNN-based edge detectors, our approach is also orders of magnitude faster. See Table~\ref{tb:bsds} for a detailed discussion.}
\label{fig:bsds}
\vspace{-1mm}
\end{figure}

We evaluate HED on the Berkeley Segmentation Dataset and Benchmark (BSDS 500)~\cite{arbelaez2011contour} which is composed of 200 training, 100 validation, and 200 testing images. Each image has manually annotated ground truth contours. Edge detection accuracy is evaluated using three standard measures: fixed contour threshold (ODS), per-image best threshold (OIS), and average precision (AP). We apply a standard non-maximal suppression technique to our edge maps to obtain thinned edges for evaluation. The results are shown in Figure~\ref{fig:bsds} and Table~\ref{tb:bsds}.

\begin{table}[!htp]
\begin{center}
\caption{Results of single and averaged side output in HED on the BSDS 500 dataset. The individual side output contributes to the fused/averaged result. Note that the learned weighted-fusion (\emph{Fusion-output}) achieves best F-score, while directly averaging all of the five layers (\emph{Average 1-5}) produces better average precision. Merging those two readily available outputs further boost the performance.}
\label{tab:hed_var}
\begin{tabular}{  m{6.6em} | m{0.8cm} | m{0.8cm} m{0.8cm} m{0.8cm}  } 
 \  & ODS & OIS & AP \\
\thickhline
Side-output 1 &  {.595} & {.620} & {.582} \\ 
Side-output 2 &  {.697} & {.715} & {.673} \\ 
Side-output 3 &  {.738} & {.756} & {.717} \\ 
Side-output 4 &  {.740} & {.759} & {.672} \\ 
Side-output 5 &  {.606} & {.611} & {.429} \\ 
\hline
\textbf{Fusion-output} & \textbf{.782} & \textbf{.802} & {.787} \\
Average 1-4            &  {.760} & {.784} & {.800} \\ 
\textbf{Average 1-5}   &  {.774} & {.797} & \textbf{.822} \\ 
Average 2-4            &  {.766} & {.788} & {.798} \\ 
Average 2-5            &  {.777} & {.800} & {.814} \\
\hline
\textbf{Merged result} & \textbf{.782} & \textbf{.804} & \textbf{.833} \\
\end{tabular}
\end{center}
\vspace{-6mm}
\end{table}

\noindent \textbf{Side outputs} To explicitly validate the side outputs, we summarize the results produced by the individual side-outputs at different scales in Table~\ref{tab:hed_var}, including different combinations of the multi-scale edge maps. We emphasize here that all the side-output predictions are obtained in one pass; this enables us to fully investigate different configurations of combining the outputs at no extra cost. There are several interesting observations from the results: for instance, combining predictions from multiple scales yields better performance; moreover, all the side-output layers contribute to the performance gain, either in F-score or averaged precision. To see this, in Table~\ref{tab:hed_var}, the side-output layer 1 and layer 5 (the lowest and highest layers) achieve similar relatively low performance. One might expect these two side-output layers to not be useful in the averaged results. However this turns out not to be the case --- for example, the Average 1-4 achieves ODS=.760 and incorporating the side-output layer 5, the averaged prediction achieves an ODS=.774. We find similar phenomenon when considering other ranges. As mentioned above, the predictions obtained using different combination strategies are complementary, and a late merging of the averaged predictions with learned fusion-layer predictions leads to the best result. Another observation is, when compared to previous "non-deep" methods, performance of all "deep" methods drops more in the high recall regime. This might indicate that deep learned features are capable of (and favor) learning the global object boundary --- thus many weak edges are omitted. HED is better than other deep learning based methods in the high recall regime because deep supervision helps us to take the low level predictions into account. 

\begin{table}[!htp]
\begin{center}
\caption{Results on BSDS500. 
 {\small $\ast$BSDS300 results,$\dag$GPU time}}
\label{tb:bsds}
\begin{tabular}{  m{7.8em} | m{7mm} m{7mm} m{7mm}| m{10mm}  } 
 \  & ODS & OIS & AP & FPS \\
\thickhline
Human & .80 & .80 & - & - \\ 
\hline
Canny & .600 & .640 & .580 & 15 \\ 
Felz-Hutt~\cite{felzenszwalb2004efficient} & .610 & .640 & .560 & 10 \\
BEL~\cite{dollar2006supervised} & .660$\ast$ & - & - & 1/10 \\
\hline
gPb-owt-ucm~\cite{arbelaez2011contour} & .726 & .757 & .696 & 1/240 \\
Sketch Tokens~\cite{lim2013sketch} & .727 & .746 & .780 & 1 \\
SCG~\cite{xiaofeng2012discriminatively} & .739 & .758 & .773 & 1/280 \\
\hline
SE-Var~\cite{Dollar2015PAMI} & .746 & .767 & .803 & 2.5 \\
OEF~\cite{hallman2014oriented} & .749 & .772 & .817 & - \\
\hline
DeepNets~\cite{kivinen2014visual} & .738 & .759 & .758 & $1/5\dag$ \\
N4-Fields~\cite{ganin2014n} & .753 & .769 & .784 & $1/6\dag$ \\
DeepEdge~\cite{bertasius2014deepedge} & .753 & .772 & .807 & 1/{$10^3$}$\dag$ \\
CSCNN~\cite{hwang2015pixel} & .756 & .775 & .798 & - \\
DeepContour~\cite{shendeepcontour} & .756 & .773 & .797 & 1/30$\dag$ \\
\hline
\textbf{HED (ours)} & \textbf{.782} & \textbf{.804} & \textbf{.833} & 2.5$\dag$, 1/12\\
\end{tabular}
\end{center}
\vspace{-5mm}
\end{table}

\noindent \textbf{Late merging to boost average precision}
We find that the weighted-fusion layer output gives best performance in F-score. However the average precision degrades compared to directly averaging all the side outputs. This might due to our focus on ``global'' object boundaries for the fusion-layer weight learning. 
Taking advantage of the readily available side outputs in HED, we merge the fusion layer output with the side outputs (at no extra cost) in order to compensate for the loss in average precision. This simple heuristic gives us the best performance across all measures that we report in Figure~\ref{fig:bsds} and Table~\ref{tb:bsds}.

\noindent \textbf{More training data} Deep models have significantly advanced results in a variety of computer vision applications, at least in part due to the availability of large training data. In edge detection, however, we are limited by the number of training images available in the existing benchmarks. Here we want to explore whether adding more training data will help further improve the results. To do this, we expand the training set by randomly sampling $100$ images from the test set. We then evaluate the result on the remaining $100$ test images. We report the averaged result over 5 such trials. 

We observe that by adding only 100 training images, performance improves from ODS=$.782$ to ODS=$.797$ ($\pm .003$), nearly touching the human benchmark. This shows a potentially promising direction to further enhance HED by training it with a larger dataset.

\subsection{NYUDv2 Dataset}
The NYU Depth (NYUD) dataset \cite{silberman2012indoor} has 1449 RGB-D images. This dataset was used for edge detection in \cite{xiaofeng2012discriminatively} and \cite{gupta2013perceptual}. Here we use the setting described in \cite{Dollar2015PAMI} and evaluate HED on data processed by \cite{gupta2013perceptual}. The NYUD dataset is split into 381 training, 414 validation, and 654 testing images. All images are made to the same size and we train our network on full resolution images.  As used in \cite{gupta2014learning,Dollar2015PAMI}, during evaluation we increase the maximum tolerance allowed for correct matches of edge predictions to ground truth from $.0075$ to $.011$.

\begin{figure}[!htp]
\begin{center}
   \includegraphics[width=0.7\linewidth]{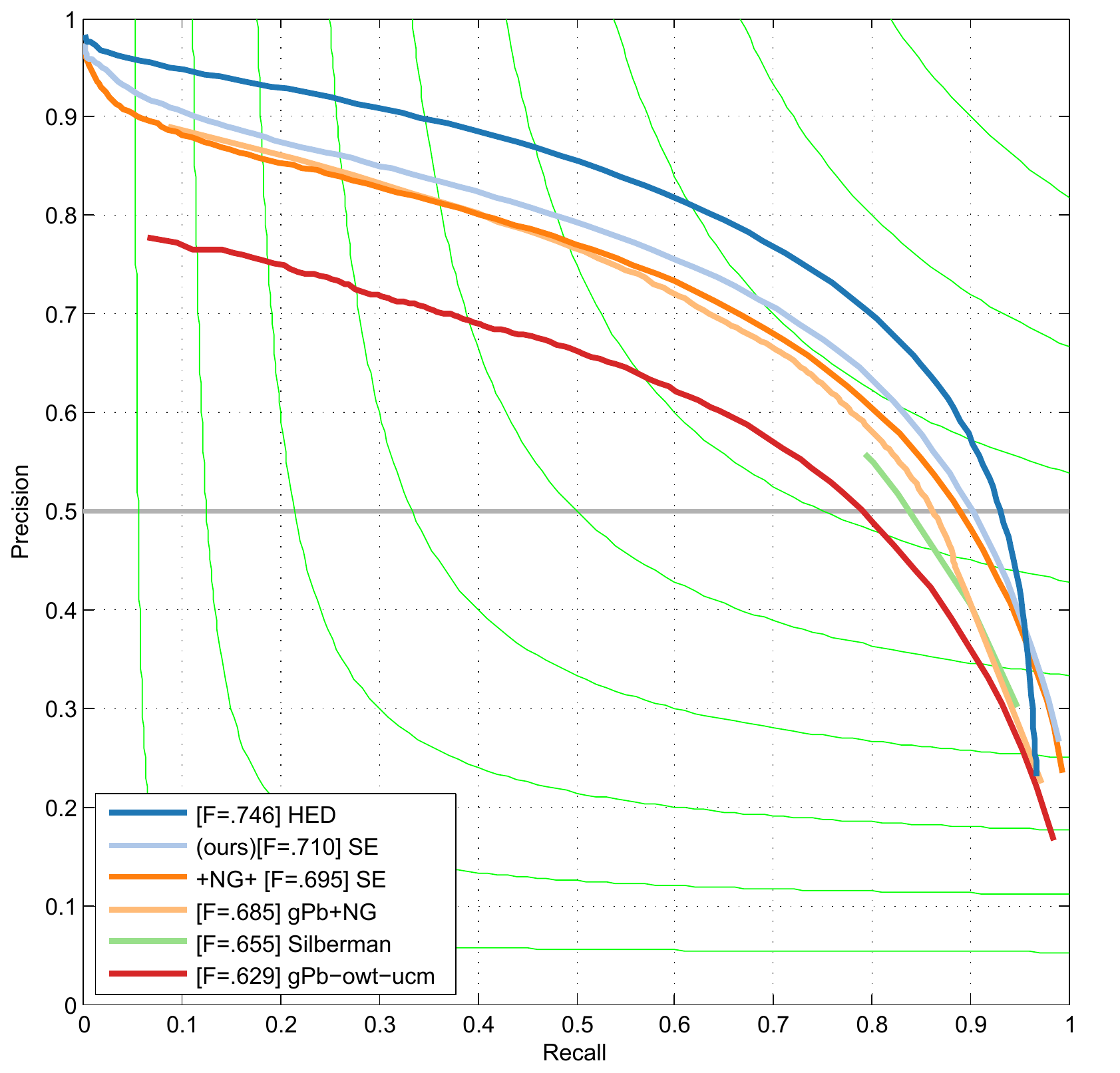}
\end{center}
\vspace{-3mm}
   \caption{\footnotesize Precision/recall curves on NYUD dataset. Holistically-nested edge detection (HED) trained with RGB and HHA features achieves the best result (ODS=.746). See Table~\ref{tab:nyud} for additional information.}
\label{fig:nyud}
\end{figure}
\vspace{-3mm}

\begin{table}[!htp]
\caption{Results on the NYUD dataset \cite{silberman2012indoor} $\dag$GPU time}
\label{tab:nyud}
\begin{center}
\begin{tabular}{  m{6.0em} | m{0.8cm} | m{0.8cm} m{0.8cm} | m{1.3cm} } 
 \  & ODS & OIS & AP & FPS\\
\thickhline
gPb-ucm &  .632 & .661 & .562 & 1/360 \\
Silberman \cite{silberman2012indoor} &  .658 & .661 & - & $<$1/360\\
gPb+NG\cite{gupta2013perceptual} & .687 & .716 & .629 & 1/375\\
\hline
SE\cite{Dollar2015PAMI} & .685 & .699 & .679  & 5\\
SE+NG+\cite{gupta2014learning} &  .710 & .723 & .738 & 1/15\\
\hline
\footnotesize{HED-RGB} &  {.720} & {.734} & {.734} & 2.5$\dag$ \\ 
\footnotesize{HED-HHA} &   {.682} & {.695} & {.702} & 2.5$\dag$ \\ 
\footnotesize{HED-RGB-HHA} & \textbf{.746} & \textbf{.761} & \textbf{.786} & 1$\dag$\\ 
\end{tabular}
\end{center}
\vspace{-10mm}
\end{table}

\noindent \textbf{Depth information encoding}
Following the success in \cite{gupta2014learning} and \cite{long2014fully}, we leverage the depth information by utilizing HHA features in which the depth information is embedded into three channels: horizontal disparity, height above ground, and angle of the local surface normal with the inferred direction of gravity . We use the same HED architecture and hyper-parameter settings as were used for BSDS 500. We train two different models in parallel, one on RGB images and another on HHA feature images, and report the results below. We directly average the RGB and HHA predictions to produce the final result by leveraging RGB-D information. We also tried other approaches to incorporate the depth information, for example, by training on the raw depth channel, or by concatenating the depth channel with the RGB channels before the first convolutional layer. None of these attempts yields notable improvement compared to the approach using HHA. The effectiveness of the HHA features shows that, although deep neural networks are capable of automatic feature learning, for depth data, carefully hand-designed features are still necessary, especially when only limited training data is available.

%\noindent \textbf{Results discussion} 
Table~\ref{tab:nyud} and Figure~\ref{fig:nyud} 
show the precision-recall evaluations of HED in comparison to other competing methods. Our network structures for training are kept the same as for BSDS. During testing we use the {\em Average2-4} prediction instead of the Fusion-layer output as it yields the best performance. We do not perform late merging since combining two sources of edge map predictions (RGB and HHA) already gives good average precision. Note that the results achieved using the RGB modality only are already better than those of the previous approaches. 

\section{Conclusion}
In this paper, we have developed a new convolutional-neural-network-based edge detection system that demonstrates state-of-the-art performance on natural images at a speed of practical relevance (e.g., $0.4$ seconds using GPU and $12$ seconds using CPU). Our algorithm builds on top of the ideas of fully convolutional neural networks and deeply-supervised nets. We also initialize our network structure and parameters by adopting a pre-trained trimmed VGGNet. Our method shows promising results %points to a promising direction
in performing image-to-image learning by combining multi-scale and multi-level visual responses, even though explicit contextual and high-level information has not been enforced. Source code and pretrained models are available online at ~\url{https://github.com/s9xie/hed}.\\

\noindent \textbf{Acknowledgment} This work is supported by NSF IIS-1216528 (IIS-1360566), NSF award IIS-0844566 (IIS-1360568), and a Northrop Grumman Contextual Robotics grant. We gratefully thank Patrick Gallagher for helping improve this manuscript. We also thank Piotr Dollar and Yin Li for insightful discussions. We are grateful for the generous donation of the GPUs by NVIDIA. 

\section*{A. More Results}
After the ICCV submission, we retrained our model with the following :  (1) In data augmentation, we further triples the dataset by scaling the training images to $50\%$, $100\%$, $150\%$ of its original size.  (2) In training phase, we use full-resolution images instead of resizing them to $400\times400$. 

\begin{figure}[!htp]
	\vspace{-3mm}
	\begin{center}
		\includegraphics[width=0.8\linewidth]{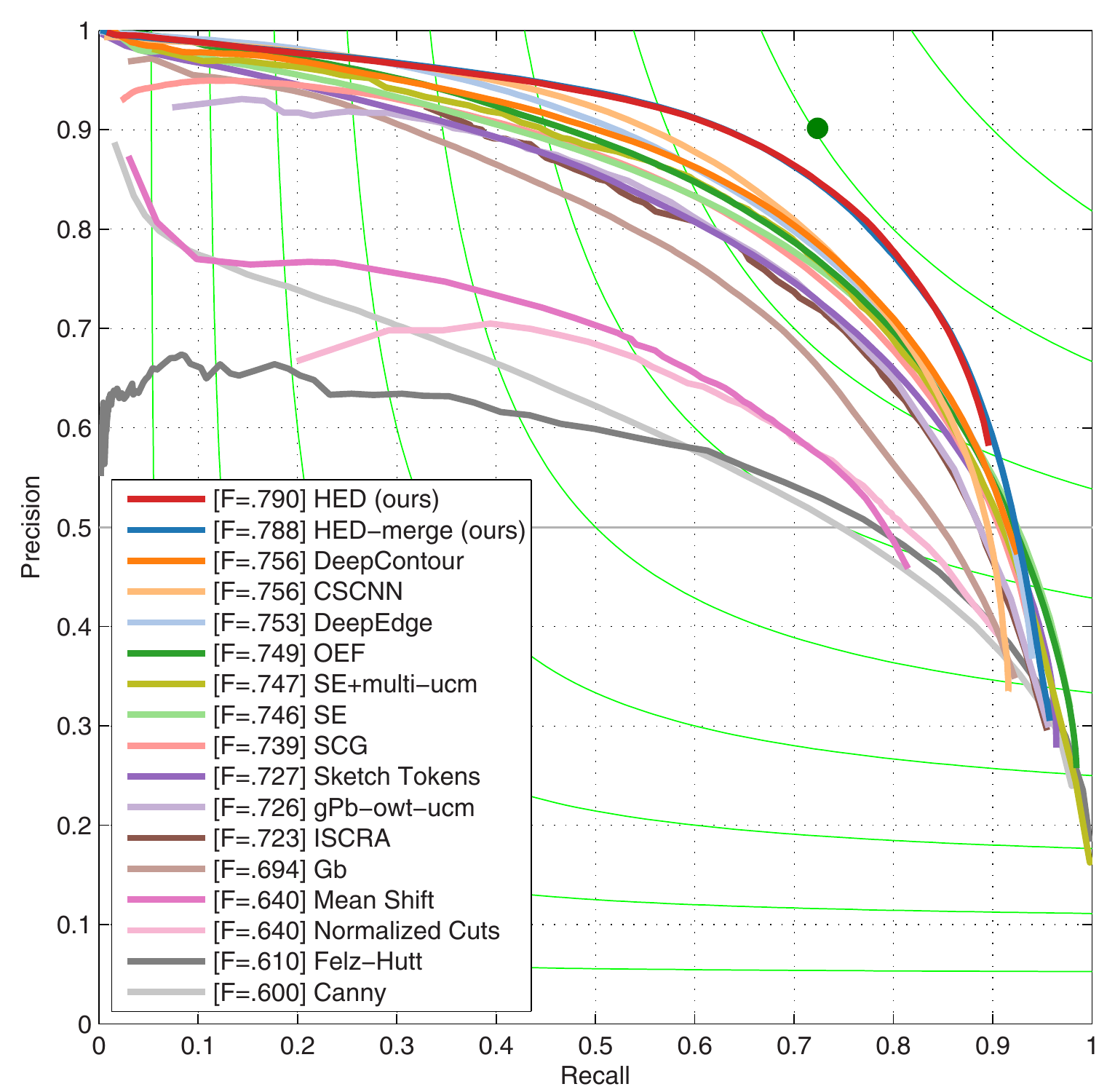}
	\end{center}
	\vspace{-4mm}
	\caption{\footnotesize Updated results on the BSDS500 dataset. Our proposed HED framework achieves the best F-score (ODS=.790, OIS=.808, AP=.811), the late-merging variant achieves best average precision (ODS=.788, OIS=.808, AP=.840).} 
	\label{fig:bsds_updated}
	\vspace{-6mm}
\end{figure}

Updated results on BSDS500 benchmark dataset with this newly trained model are reported in Figure~\ref{fig:bsds_updated} and Table~\ref{tab:bsds_updated}.

In the new experiment settings, while we found that the gap in F-score narrows between models with/without deep supervision, we have similar qualitative and quantitative observations as illustrated in Section~\ref{sec:alt}.

\begin{table}[!htp]
	\begin{center}
		\caption{Updated HED results on the BSDS500 dataset.}
		\label{tab:bsds_updated}
		\begin{tabular}{  m{12.8em} | m{0.4cm} m{0.4cm} m{0.4cm}  } 
			\  & ODS & OIS & AP \\
			\thickhline
			\footnotesize{\textbf{fusion-output (with deep supervision)}} & \textbf{.790} & \textbf{.808} & {.811} \\
			\footnotesize{\textbf{fusion-output (w/o deep supervision)}} & .785 & .801 & .730 \\
			\footnotesize{\textbf{HED (late-merging)}} & .788 & .808 & \textbf{.840} \\
		\end{tabular}
	\end{center}
	\vspace{-6mm}
\end{table}

\subsection*{Changelog}
\textbf{v2} Fix typos and reorganize formulations. Add Table~\ref{tab:hed_fcn} to discuss the role of deep supervision. Add appendix A for updated results on BSDS500 in a new experiment setting. Add links to publicly available repository for training/testing code, augmented data and pre-trained model.

{\small
\bibliographystyle{ieee}
\bibliography{egbib}

\begin{thebibliography}{10}\itemsep=-1pt

\bibitem{arbelaez2011contour}
P.~Arbelaez, M.~Maire, C.~Fowlkes, and J.~Malik.
\newblock Contour detection and hierarchical image segmentation.
\newblock {\em PAMI}, 33(5):898--916, 2011.

\bibitem{bertasius2014deepedge}
G.~Bertasius, J.~Shi, and L.~Torresani.
\newblock Deepedge: A multi-scale bifurcated deep network for top-down contour
  detection.
\newblock In {\em CVPR}, 2015.

\bibitem{buyssens2013multiscale}
P.~Buyssens, A.~Elmoataz, and O.~L{\'e}zoray.
\newblock Multiscale convolutional neural networks for vision--based
  classification of cells.
\newblock In {\em ACCV}. 2013.

\bibitem{canny1986computational}
J.~Canny.
\newblock A computational approach to edge detection.
\newblock {\em PAMI}, (6):679--698, 1986.

\bibitem{dollar2006supervised}
P.~Dollar, Z.~Tu, and S.~Belongie.
\newblock Supervised learning of edges and object boundaries.
\newblock In {\em CVPR}, 2006.

\bibitem{Dollar2015PAMI}
P.~Doll\'ar and C.~L. Zitnick.
\newblock Fast edge detection using structured forests.
\newblock {\em PAMI}, 2015.

\bibitem{elder2002ecological}
J.~H. Elder and R.~M. Goldberg.
\newblock Ecological statistics of gestalt laws for the perceptual organization
  of contours.
\newblock {\em Journal of Vision}, 2(4):5, 2002.

\bibitem{farabet2013learning}
C.~Farabet, C.~Couprie, L.~Najman, and Y.~LeCun.
\newblock Learning hierarchical features for scene labeling.
\newblock {\em PAMI}, 2013.

\bibitem{felzenszwalb2004efficient}
P.~F. Felzenszwalb and D.~P. Huttenlocher.
\newblock Efficient graph-based image segmentation.
\newblock {\em IJCV}, 59(2):167--181, 2004.

\bibitem{ganin2014n}
Y.~Ganin and V.~Lempitsky.
\newblock N4-fields: Neural network nearest neighbor fields for image
  transforms.
\newblock {\em arXiv preprint arXiv:1406.6558}, 2014.

\bibitem{gupta2013perceptual}
S.~Gupta, P.~Arbelaez, and J.~Malik.
\newblock Perceptual organization and recognition of indoor scenes from rgb-d
  images.
\newblock In {\em CVPR}, 2013.

\bibitem{gupta2014learning}
S.~Gupta, R.~Girshick, P.~Arbel{\'a}ez, and J.~Malik.
\newblock Learning rich features from rgb-d images for object detection and
  segmentation.
\newblock In {\em ECCV}, 2014.

\bibitem{hallman2014oriented}
S.~Hallman and C.~C. Fowlkes.
\newblock Oriented edge forests for boundary detection.
\newblock {\em arXiv preprint arXiv:1412.4181}, 2014.

\bibitem{hariharan2014hypercolumns}
B.~Hariharan, P.~Arbel{\'a}ez, R.~Girshick, and J.~Malik.
\newblock Hypercolumns for object segmentation and fine-grained localization.
\newblock In {\em CVPR}, 2015.

\bibitem{hoiem2008putting}
D.~Hoiem, A.~A. Efros, and M.~Hebert.
\newblock Putting objects in perspective.
\newblock {\em IJCV}, 80(1):3--15, 2008.

\bibitem{hoiem2007recovering}
D.~Hoiem, A.~N. Stein, A.~A. Efros, and M.~Hebert.
\newblock Recovering occlusion boundaries from a single image.
\newblock In {\em ICCV}, 2007.

\bibitem{hou2013boundary}
X.~Hou, A.~Yuille, and C.~Koch.
\newblock Boundary detection benchmarking: Beyond f-measures.
\newblock In {\em CVPR}, 2013.

\bibitem{hubel1962receptive}
D.~H. Hubel and T.~N. Wiesel.
\newblock Receptive fields, binocular interaction and functional architecture
  in the cat's visual cortex.
\newblock {\em The Journal of physiology}, 160(1):106--154, 1962.

\bibitem{hwang2015pixel}
J.-J. Hwang and T.-L. Liu.
\newblock Pixel-wise deep learning for contour detection.
\newblock In {\em ICLR}, 2015.

\bibitem{kittler1983accuracy}
J.~Kittler.
\newblock On the accuracy of the sobel edge detector.
\newblock {\em Image and Vision Computing}, 1(1):37--42, 1983.

\bibitem{kivinen2014visual}
J.~J. Kivinen, C.~K. Williams, N.~Heess, and D.~Technologies.
\newblock Visual boundary prediction: A deep neural prediction network and
  quality dissection.
\newblock In {\em AISTATS}, 2014.

\bibitem{konishi2003statistical}
S.~Konishi, A.~L. Yuille, J.~M. Coughlan, and S.~C. Zhu.
\newblock Statistical edge detection: Learning and evaluating edge cues.
\newblock {\em PAMI}, 25(1):57--74, 2003.

\bibitem{DSN}
C.-Y. Lee, S.~Xie, P.~Gallagher, Z.~Zhang, and Z.~Tu.
\newblock Deeply-supervised nets.
\newblock In {\em AISTATS}, 2015.

\bibitem{lim2013sketch}
J.~J. Lim, C.~L. Zitnick, and P.~Doll{\'a}r.
\newblock Sketch tokens: A learned mid-level representation for contour and
  object detection.
\newblock In {\em CVPR}, 2013.

\bibitem{liu2011nonparametric}
C.~Liu, J.~Yuen, and A.~Torralba.
\newblock Nonparametric scene parsing via label transfer.
\newblock {\em PAMI}, 33(12):2368--2382, 2011.

\bibitem{long2014fully}
J.~Long, E.~Shelhamer, and T.~Darrell.
\newblock Fully convolutional networks for semantic segmentation.
\newblock In {\em CVPR}, 2015.

\bibitem{marr1980theory}
D.~Marr and E.~Hildreth.
\newblock Theory of edge detection.
\newblock {\em Proceedings of the Royal Society of London. Series B. Biological
  Sciences}, 207(1167):187--217, 1980.

\bibitem{martin2004learning}
D.~R. Martin, C.~C. Fowlkes, and J.~Malik.
\newblock Learning to detect natural image boundaries using local brightness,
  color, and texture cues.
\newblock {\em PAMI}, 26(5):530--549, 2004.

\bibitem{neverova2014multi}
N.~Neverova, C.~Wolf, G.~W. Taylor, and F.~Nebout.
\newblock Multi-scale deep learning for gesture detection and localization.
\newblock In {\em ECCV Workshops}, 2014.

\bibitem{ren2008multi}
X.~Ren.
\newblock Multi-scale improves boundary detection in natural images.
\newblock In {\em ECCV}. 2008.

\bibitem{xiaofeng2012discriminatively}
X.~Ren and L.~Bo.
\newblock Discriminatively trained sparse code gradients for contour detection.
\newblock In {\em NIPS}, 2012.

\bibitem{ruderman1994statistics}
D.~L. Ruderman and W.~Bialek.
\newblock Statistics of natural images: Scaling in the woods.
\newblock {\em Physical review letters}, 73(6):814, 1994.

\bibitem{sermanet2012convolutional}
P.~Sermanet, S.~Chintala, and Y.~LeCun.
\newblock Convolutional neural networks applied to house numbers digit
  classification.
\newblock In {\em ICPR}, 2012.

\bibitem{shendeepcontour}
W.~Shen, X.~Wang, Y.~Wang, X.~Bai, and Z.~Zhang.
\newblock Deepcontour: A deep convolutional feature learned by positive-sharing
  loss for contour detection draft version.
\newblock In {\em CVPR}, 2015.

\bibitem{silberman2012indoor}
N.~Silberman, D.~Hoiem, P.~Kohli, and R.~Fergus.
\newblock Indoor segmentation and support inference from rgbd images.
\newblock In {\em ECCV}. 2012.

\bibitem{vgg}
K.~Simonyan and A.~Zisserman.
\newblock Very deep convolutional networks for large-scale image recognition.
\newblock In {\em ICLR}, 2015.

\bibitem{torre1986edge}
V.~Torre and T.~A. Poggio.
\newblock On edge detection.
\newblock {\em PAMI}, (2):147--163, 1986.

\bibitem{tu2008auto}
Z.~Tu.
\newblock Auto-context and its application to high-level vision tasks.
\newblock In {\em CVPR}, 2008.

\bibitem{van1994neural}
D.~C. Van~Essen and J.~L. Gallant.
\newblock Neural mechanisms of form and motion processing in the primate visual
  system.
\newblock {\em Neuron}, 13(1):1--10, 1994.

\bibitem{witkin1984scale}
A.~P. Witkin.
\newblock Scale-space filtering: A new approach to multi-scale description.
\newblock In {\em ICASSP}, 1984.

\bibitem{yuille1986scaling}
A.~L. Yuille and T.~A. Poggio.
\newblock Scaling theorems for zero crossings.
\newblock {\em PAMI}, (1):15--25, 1986.

\end{thebibliography}
}

\end{document}